\documentclass[10pt,twocolumn,letterpaper]{article}

\usepackage{iccv}              % To produce the CAMERA-READY version
\usepackage{algorithmic}
\usepackage{amsmath}
\usepackage{multirow}
\usepackage{xcolor}
\usepackage{enumitem}
\usepackage{colortbl}
\usepackage[linesnumbered,ruled,vlined, noend]{algorithm2e}

\SetCommentSty{mycommfont}

\usepackage{caption}
\usepackage{xspace}

\definecolor{verylightblue}{RGB}{220,235,245}
\newcommand{\baseline}[1]{\cellcolor{verylightblue}{#1}}

\newcommand{\sys}{\textsc{StreamMind}\xspace}

\definecolor{iccvblue}{rgb}{0.21,0.49,0.74}
\usepackage[pagebackref,breaklinks,colorlinks,allcolors=iccvblue]{hyperref}

\title{\sys: Unlocking Full Frame Rate Streaming Video Dialogue through Event-Gated Cognition}

\author{%
  \textbf{Xin Ding}\textsuperscript{1}\hspace{4pt}\thanks{This work was done during Xin Ding's internship at Microsoft Research.}
  \quad
  \textbf{ Hao Wu}\textsuperscript{3}\hspace{4pt}\textsuperscript{$\dag$}
  \quad
  \textbf{ Yifan Yang}\textsuperscript{2}
  \quad
  \textbf{ Shiqi Jiang}\textsuperscript{2}
  \quad
  \textbf{ Qianxi Zhang}\textsuperscript{2}\\
  \quad
  \textbf{ Donglin Bai}\textsuperscript{2}
  \quad
  \textbf{  Zhibo Chen}\textsuperscript{1} 
  \quad
  \textbf{ Ting Cao}\textsuperscript{4}\hspace{4pt}\textsuperscript{$\dag$} \\
  \textsuperscript{1}University of Science and Technology of China 
  \quad
  \textsuperscript{2}Microsoft Research
  \quad
  \textsuperscript{3} Nanjing University\quad \\
  \textsuperscript{4} Institute for AI Industry Research (AIR), Tsinghua University
  \\
}
\begin{document}
\maketitle

\begin{abstract}
With the rise of real-world human-AI interaction applications, such as AI assistants, the need for Streaming Video Dialogue is critical. To address this need, we introduce \sys, a video LLM framework that achieves ultra-FPS streaming video processing (100\,fps on a single A100) and enables proactive, always-on responses in real time, without explicit user intervention. 

To solve the key challenge of the contradiction between linear video streaming speed and quadratic transformer computation cost, we propose a novel perception-cognition interleaving paradigm named \textbf{``event-gated LLM invocation''}, in contrast to the existing per-time-step LLM invocation. By introducing a \textbf{Cognition Gate} network between the video encoder and the LLM, LLM is only invoked when relevant events occur. To realize the event feature extraction with constant cost, we propose Event-Preserving Feature Extractor (EPFE) based on state-space method, generating a single perception token for spatiotemporal features. These techniques enable the video LLM with full-FPS perception and real-time cognition response. 

Experiments on Ego4D and SoccerNet streaming tasks, as well as standard offline benchmarks, demonstrate state-of-the-art performance in both model capability and real-time efficiency,  paving the way for ultra-high-FPS applications, such as Game AI  and interactive media.   The code and data is available at \href{https://aka.ms/StreamMind}{https://aka.ms/StreamMind}.

\end{abstract}
\section{Introduction}
The advancement of large foundation models is driving an increasing number of real-world human-AI interaction applications, such as AI home companions~\cite{merrill2022ai,battistoni2023using} and human-robot collaboration~\cite{vysocky2016human,xu2024pllava}. These applications heavily rely on streaming video understanding capabilities~\cite{zhang2024flash,qian2025streaming,distreaming,xiong2025streaming,huang2024online}, specifically streaming video dialogue~\cite{chen2024videollm,wu2025videollm}. Streaming video dialogue (StreamingVD) aims to continuously perceive incoming video streams and, based on user queries, provide proactive, real-time, and always-on responses without human intervention. For example, an AI companion could respond to user queries like ``Guide me to fix this bicycle'' or ``Let's watch and discuss the soccer game together''.
%More and more real-world applications like AI companions, assistants, robots, and live broadcasts require streaming video understanding capability. It can operate the streaming modality input and provide online timely understanding and reasoning to fulfill tasks. This sparks the research interests in streaming video QA, narration, and . }

\begin{figure*}[!th] %H为当前位置，!htb为忽略美学标准，htbp为浮动图形
\centering %图片居中
\includegraphics[width=0.8\textwidth]{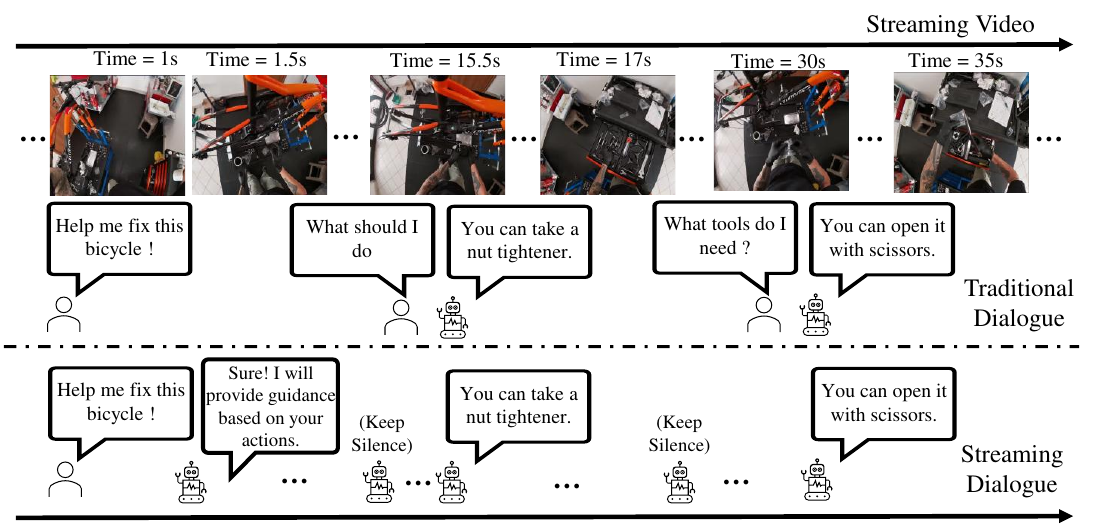} %插入图片，[]中设置图片大小，{}中是图片文件名
\caption{Comparison between streaming dialogue (bottom) and traditional dialogue tasks (top). Streaming dialogue proactively generates responses at appropriate time based on user queries and perception, whereas traditional dialogue requires human triggering for each response. %StreamingVD enables smarter interactive scenarios, requiring the ability to proactively respond at the future appropriate moments, as demonstrated in the bottom examples. 
}
\label{fig:scenario} %用于文内引用的标签
\end{figure*}

Streaming video dialogue can be viewed as a generalization of standard offline video dialogue tasks~\cite{yang2022zero,maaz2023video,ren2024timechat,li2024llama,li2024mvbench,luo2023valley,yang2023vid2seq}, extending video LLMs to broader real-world scenarios. The fundamental challenge of StreamingVD compared to its offline counterpart lies in the requirement for \textbf{timing alignment between events and responses}. That is, as the video progresses, the model is required to \textbf{(1) proactively decide when to respond and (2) respond in real time before the next event occurs} according to user queries, as illustrated in Fig.~\ref{fig:scenario}. However, with current techniques, these two requirements are contradictory and cannot be achieved simultaneously.
\begin{figure}[t] %H为当前位置，!htb为忽略美学标准，htbp为浮动图形
\centering %图片居中
\includegraphics[width=0.5\textwidth]{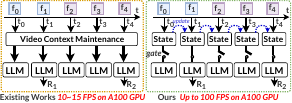} %插入图片，[]中设置图片大小，{}中是图片文件名
\caption{The paradigm of existing per-step LLM invocation (left) versus our event-gated LLM invocation (right) for streaming video dialogue task. Our paradigm boosts frame rate up-to 100\,fps on A100. ($R_i$: response, $f_i$: frame, t: time) }%Model Processing Speed for Streaming Video by $10\times$.} %最终文档中希望显示的图片标题
\label{fig:paradigm} %用于文内引用的标签
\vspace{-5mm}
\end{figure}

%However, existing works~\cite{} cannot achieve the two requirements simultaneously, due to the much slower processing speed compared to the video stream rate. 
The seminal VideoLLM-Online~\cite{chen2024videollm} and VideoLLM-MoD~\cite{wu2025videollm} introduce a \textbf{per-step LLM invocation} paradigm for StreamingVD, shown in Fig.~\ref{fig:paradigm}. At each time step, it inputs all past frames and the user query to the LLM, to decide when to respond or remain silent. However, the $O(n^3)$\footnote{Invoking a quadratic-complexity LLM n times results in cubic complexity.} computational complexity and limited context window are mismatched with the $O(n)$ frame streaming, making real-time response difficult to achieve. More works~\cite{distreaming,qian2025streaming,xiong2025streaming,zhang2024flash} focus on enhancing the efficiency of offline video dialogue aiming to meet real-time requirements. However, similar to offline approaches, they depend on user queries to trigger responses rather than the model to do so proactively. 

To empower StreamingVD, this paper proposes a new perception-cognition interleaving paradigm, named \textbf{event-gated LLM invocation}, shown in Fig.~\ref{fig:paradigm}. It is inspired by the \textit{event perception mechanism} of human brains~\cite{zacks2007event}, %where people perceive activities as discrete events
which states that events are key components of
human's perception, attention, and memory\footnote{A formal definition of an event~\cite{zacks2007event}: “A segment of time at a given location that is conceived by an observer to have a beginning and an end."}. %That is to say, humans tend to put attention only on events. 
We therefore propose an event-based selective cognition process. Rather than invoking the LLM at every time step, we introduce a \textbf{Cognition Gate} between the video encoder and LLM. The video encoder continuously perceives visual signals of each frame, and only when query-related events occur, the gate will open to invoke the LLM. Analogous to how a text prompt triggers an LLM, streaming visual signals lack a clear stopping point, unlike a finite text sequence. The cognition gate bridges this gap by acting as a mechanism to identify ``physical world prompts'' for the LLM. %ensuring that cognition is triggered only when necessary.}

This new paradigm raises two major technical concerns, to achieve both high capability and efficiency. Firstly, existing video encoders in VideoLLMs (e.g., STC in VideoLLaMA) only capture local spatio-temporal features. This results in a linear increase in the number of feature tokens for streaming frames and subsequently a quadratic computation for the processing, failing to meet real-time requirements.
%\xin{Firstly, with continuous streaming frames input, existing video encoders in VideoLLMs (e.g., STC in VideoLLaMA) only capture local spatio-temporal features, leading to an $O(n)$ feature tokens and subsequently an $O(n^2)$ computational complexity for the processing, failing to meet efficiency requirements.}
% Firstly, the video encoder needs to capture event-level spatio-temporal features. However, existing video encoders in video LLMs (e.g., STC in videoLLama~\cite{}) only capture local spatio-temporal features, relying on the quadratic attention computation for long-term dependencies. This again is mismatched with the streaming speed.
%the input to the gate from the video encoder needs to extract event features (could last hundreds of frames) in real time. However, general video encoders in video LLMs, such as the STC in Video Llama, only capture local spatio-temporal features (dozens of frames) and rely on the quadratic attention for long-term dependencies.   \xin{Firstly, existing video encoders in videoLLMs (e.g. STC in videoLLama) only capture local spatio-temporal features, resulting in the cognition gate requiring $O(n)$ feature tokens as input and incurring $O(n^2)$ perceptual computation complexity, which fails to meet efficiency requirements.}
%Secondly, the Gate needs to decide in real time whether to invoke the LLM based on visual features and the user query. However, simply attending visual and query tokens  could limit the gate capability to feature-matching and retrieval-like tasks. 
Secondly, the Cognition Gate must decide to invoke the LLM or not, based on visual features and the user query. However, only considering efficiency and simply attending visual and query tokens (e.g., Cross Attention in Q-former~\cite{li2023blip})  limits the gate capability to tasks like feature-matching~\cite{wang2023instructta,wang2022matchformer} and retrieval~\cite{zhang2025tpcap,amoroso2024perceive,fang2024enhancing}, struggle to make decisions with deeper semantic understanding. 

%\ting{[Our techniques] To address the challenges, we realize \sys, a novel framework for real streaming video understanding, with the processing speed matching frame streaming speed, and \ting{no limitation on context window? or process hours-long videos}. It integrates two techniques. For the video encoder, we utilize the state-assisted module to extract the temporal features, for its constant computing and space cost as well as the suitability for modeling continuous physical signals. Its SOTA structured state space model also allows for long-term information retention. For the Cognition Gate, to leverage the reasoning capability and world knowledge of the LLM as well as meeting the real-time requirement, the Gate is designed to reuse the shallow layers of the LLM. Since the gating decision is binary (Yes/No), deeper layers—primarily responsible for nuanced language generation—are unnecessary.}

By addressing the challenges, we realize \sys, a novel framework for  streamingVD, with the processing speed matching frame rate for the first time.  %\ting{no limitation on context window? or process hours-long videos}. 
It integrates two techniques. For the video encoder,  inspired by the  exceptional ability of the state-assisted module~\cite{mamba} in modeling  continuous physical signals with constant cost, we propose the \textit{Event-Preserving Feature Extractor} (EPFE) based on the state-assisted module. It generates a single perception token of event features to the gate, resulting in a constant perception cost. % reducing the computational complexity of the perception process to $O(1)$. 
For the Cognition Gate, to leverage the world knowledge of LLM while meeting the real-time requirement, we propose a \textit{Shallow Layer Transfer} method. The Gate is designed to reuse the shallow layers of the LLM and its autoregressive training to maximize the probabilities of ``$<response/silence>$'' tokens during training.%Since the gating decision is binary (Yes/No), deeper layers—primarily responsible for nuanced language generation—are unnecessary.

% \ting{- Achieving ultra frames paving the way for real streaming physical world understanding for embodied AI, movies, game, navigating ..}
To evaluate the performance of the \sys framework, we adopt evaluation metrics from VideoLLM-Online and a wide range of text generation evaluation metrics. Additionally, we design two novel metrics to comprehensively assess the model’s temporal responsiveness. Experiments on real-time Ego4D~ \cite{grauman2022ego4d} and SoccerNet~\cite{giancola2018soccernet} demonstrate superior performance across all metrics. Furthermore, our model achieves state-of-the-art results on multiple offline benchmarks, including short- and long-term activity recognition and forecasting on the COIN~\cite{tang2019coin} and Ego4D LTA~\cite{huang2023palm} datasets. Notably, our framework breaks the ultra-FPS streaming video processing bottleneck, enabling streamingVD at up to 100\,fps on a single A100 GPU, laying a robust foundation for real-world human-AI interaction applications.

\section{Method}

\subsection{Task Discussion and Definition}   

\paragraph{Problem formulation of StreamingVD} %Given a video stream $\mathcal{V}^T:= [v_1, v_2 ,..., v_T]$ consisting of $T$ frames and a set of queries $\mathcal{Q}:= \left \{q_1, q_2,..., q_N  \right \}$, traditional video dialogue focuses solely on answering a given question $q_i$ based on events occurring up to time $t (1 < t \le T)$, regardless of whether the setting is online or offline. But StreamingVD is also tasked with processing a query at any time step $t_s (1\le t_s\le T)$ and generating real-time responses at appropriate future moments until termination at $t_e (t_s < t_e \le T)$. In this process, the AI assistant maintains a continuous real-time comprehension of the streaming video at every time step $t_i (t_s \le t_i \le t_e)$, which can be formulated as:
Given a video stream $\mathcal{V}^T:= [v_1, v_2 ,..., v_T]$ consisting of $T$ frames and a set of queries $\mathcal{Q}:= \left \{q_1, q_2,..., q_N  \right \}$. Traditional video dialogue, shown in Fig.~\ref{fig:scenario}, provides a response triggered by a user query. For a query $q_i$ given at time step $t_s$, it responds based on past frames up to time $t_s (1 < t_s \le T)$, regardless of an online or offline scenario. On the other hand, StreamingVD proactively generates responses to a query given at $t_s$ at any appropriate time steps until the query termination at $t_e (t_s < t_e \le T)$. At each time step $t_i  (t_s \le t_i \le t_e)$, StreaminVD needs to decide whether to generate a response according to the past frames and query, and then generate the response, formulated as follows.  
\begin{equation} \max P(\texttt{[Res$^{t_i}$]} \mid \texttt{[Ctx$^{<t_i}$]}, \texttt{[F$^{t_i}$]}) \end{equation}
where $\texttt{[Res$^{t_i}$]} \in \{\texttt{[Txt$^{t_i}$]}, \texttt{[EOS$^{t_i}$]}\}$ means generate a response at a time step $t_i$, or remain silence, i.e., \texttt{[EOS]}. $\texttt{[Ctx$^{<t_i}$]}$ denotes the contextual tokens accumulated from past frames and queries before $t_i$.  $\texttt{[F$^{t_i}$]}$ represents the visual features extracted from the video frame at $t_i$. 

\paragraph{Challenges of StreamingVD tasks} Firstly, increasing number of frames versus real-time requirement. Different from the offline tasks\cite{xu2017video,yu2019activitynet,li2024mvbench,fang2025mmbench,fu2024video}, streaming tasks\cite{chen2024videollm,huang2024online,qian2025streaming,xiong2025streaming,distreaming,zhang2024flash} require time alignment between event and response. Besides only accuracy constraint of offline task, streaming video tasks need to consider both \textbf{accuracy and timeliness constraints}. Offline tasks can trade resources and latency for better quality, while streaming task cannot. 

Secondly, proactive response versus real-time requirement. Traditional video dialogue~\cite{yang2021just,xiao2021next,di2024grounded,qian2025streaming,xiong2025streaming,distreaming}  are reactive, responding only at the time of user query. However, real-world dialogue is often without human triggering. The user may lack the specific knowledge to formulate questions. Similarly, embodied AI should anticipate user needs rather than merely reacting to commands.  However, proactive response means at every frame the model needs to judge, inceasing the model cost by O(n) times.       

% \paragraph{Discussion: Core Capabilities of StreamingVD}
% % Traditional video dialogue just focuses on callback an incoming video stream and providing an answers to user questions. In contrast, StreamingVQN requires the AI assistant to proactively trigger when to respond at any future moment after the requirement is posed, while also generating real-time responses at these moments. Thus, Streaming VQN not only requires the real-time capability of StreamingVQA but also the ability to proactively trigger responses. 
% \xin{Traditional video dialogue~\cite{yang2021just,xiao2021next,di2024grounded,qian2025streaming,xiong2025streaming,distreaming} focuses solely on responding to user queries based on the current video information, whether in an online or offline manner,as shown in fig.\ref{fig:scenario}. However, the introduction of streaming video increases the demand for handling future events. Therefore, StreamingVD~\cite{wu2025videollm,chen2024videollm} not only fulfills the tasks of traditional video dialogue but also requires the AI assistant to proactively determine when to respond at any future moment after the query is posed while generating real-time responses at these moments.}

\paragraph{Existing works} Conventional offline Video LLMs \cite{lin2023video,li2024llama,lin2024vila,chen2024internvl,zhang2024direct,zhang2024long} typically employ a visual encoder~\cite{radford2021learning,zhai2023sigmoid,fang2023eva} and a projection module~\cite{li2023blip,luo2023valley} to process uniformly sampled frames. The output is concatenated with the tokenized question to form a sequence, which is then passed to an LLM decoder to predict a response. However, this approach is impractical for streaming tasks due to the high computational cost associated with processing a large number of frames ($T$). Therefore, works such as \cite{xiong2025streaming,distreaming} have implemented a sliding-window attention mechanism and a KV-cache system to reduce computational overhead, or incorporated a parallel sampling strategy to enhance processing speed and reduce latency. However, these methods do not proactively trigger response when needed. In contrast, Videollm-Online \cite{chen2024videollm} and MOD \cite{wu2025videollm} utilize the LLM decoder to process every frame of streaming videos, resulting in a cubic computation cost. %which limits their capability to handle videos at only 10 fps. 
\sys, on the other hand, has the potential to process streaming videos at 100 fps, which is crucial for enabling interactive video analysis applications.

\begin{figure*}[h] %H为当前位置，!htb为忽略美学标准，htbp为浮动图形
\centering %图片居中
\includegraphics[width=0.95\textwidth]{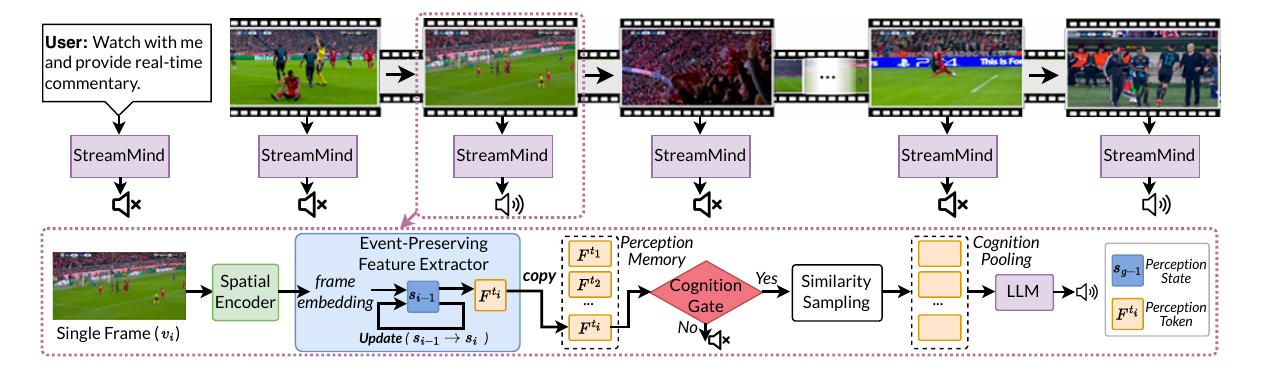} %插入图片，[]中设置图片大小，{}中是图片文件名
\caption{\sys workflow. For each video frame, the perception stage goes through Clip for spatial features and the proposed EPFE for spatiotemporal features, to generate a single perception token saved in perception memory. The Cognition Gate determines whether to invoke LLM only based on the current perception token. If yes (a query-related event occurs), the tokens in perception memory are sampled to cognition pooling as the input for LLM for cognition response. The gate decouples perception and cognition to enable full-FPS perception and event-based cognition.}%Workflow of the proposed \sys. In the perception stage, we introduce Cognition Gate, a lightweight model that accepts user requirements and processes each incoming frame. As the video stream progresses, we only need to handle a single perception token at a time to capture ongoing events. Once the user’s requirements are met, the Cognition Gate proactively triggers the cognition stage. In this stage, the LLM is engaged only once to perform high-level reasoning and generate insightful, context-aware responses.}
% \sys introduces a dual-memory mechanism, consisting of immediate memory $\mathcal{M}_\text{immediate}$ and historical synopsis $\mathcal{M}_\text{historical}$.A Temporal-Aware Feature Extractor processes video frames in real-time, constructing and maintaining memory information that preserves the temporal characteristics of the streaming video.  The Cognition Gate continuously assesses whether the accumulated Immediate Memory contains sufficient information, determining whether to engage the LLM for further processing and update the Historical Synopsis accordingly.\ting{this figure should be improveds. It did not show your advantage clearly.}} %最终文档中希望显示的图片标题
\label{fig:framework} %用于文内引用的标签
\end{figure*}

\subsection{The Overview of \sys}

For a streaming video sequence $[v^{t_s \leq t_i \leq t_e}]$, \sys models the following process as shown in Fig.~\ref{fig:framework}:

\textbf{1. Perception Phase:} With continuous streaming video, each incoming frame 
$v$ is processed through the CLIP~\cite{radford2021learning} encoder to extract spatial features. The features are then input to our proposed EPFE. EPFE updates its hidden state to correlate the frame's features across temporal dimensions and reduce redundant spatial information.% that is irrelevant to the ongoing event. 
This process generates a new perception token $\texttt{[F}^{t_i}_{per}\texttt{]}$ and saves this token in the Perception Memory $ \mathcal{M}_\text{per}^{t_i}$. The perception phase can be formulated as follows:
\begin{equation}
\label{equ:perception token}
    \texttt{[F}^{t_i}_{per}\texttt{]} = \texttt{EPFE}(\texttt{CLIP}(v^{t_i}), \texttt{H}^{t_i-1})
\end{equation}
where  $\texttt{H}^{t-1}$ denotes the hidden state from the previous time.

\textbf{2. Cognition Gate Judgment:} We introduce Cognition Gate $\mathcal{G}$ between perception and cognition. The input are user query $\texttt{[Prompt]}$ and the generated perception token $\texttt{[F}^{t_i}_{per}\texttt{]}$ at each time step $t_i$.  The Cognition Gate learns to determine whether to trigger the cognition phase at $t_i$: 
\begin{equation} 
\label{equ:cognition gate}
\max P(\texttt{[Res}^{t_i}\texttt{]} \mid \mathcal{G}(\texttt{[Prompt]}, \texttt{[F}^{t_i}_{per}\texttt{]})) \end{equation}
where $\texttt{[Res}^{t_i}\texttt{]}\in \{</response>,</silence>\}$ determines whether to trigger a response or remain silent.

\textbf{3. Cognition Phase} Upon receiving the trigger signal from the Cognition Gate,  \sys samples perception tokens from the Perception Memory $\mathcal{M}_\text{per}^{t_i}$ to the Cognition Pooling as the input to LLM to generate response.

\subsection{Perception Phase}

\textbf{Motivation:} 
To match the constant frame rate of streaming videos, ideally the perception phase should be constant computation cost for processing each frame. However, current video encoders generally extract local (dozens of frames) spatiotemporal features with constant cost,  and input all the generated perception tokens to a transformer model to extract the long-term temporal relationships. The computational cost is $O(n^3)$, obviously the lagger behind the streaming rate.

To solve the issue, we propose Event-Preserving Feature Extractor (EPFE) to replace the projection module used in existing video LLMs (e.g., cross-attention and Q-Former). EPFE is based on the SOTA state-space model, i.e., Selective State Space model~\cite{mamba}. As shown in Fig.~\ref{fig:framework} (blue box) and Eq.~\ref{eq:state_space}, for each frame, EPFE dynamically correlates the spatial features and the past internal states to generate a single perception token $\texttt{[F}^{t_i}_{per}\texttt{]}$ as well as update its state. This enables a matched speed of streaming video and perception processing, for full FPS StreamingVD. 
\begin{equation}
    \mathbf{h}_{t+1} = \mathbf{A} \mathbf{h}_t + \mathbf{B} \mathbf{x}_t, \quad 
    \mathbf{y}_t = \mathbf{C} \mathbf{h}_t
\label{eq:state_space}
\end{equation}
where $\mathbf{A},\mathbf{B}$, and $\mathbf{C}$ are learnable state-space matrices for state, input, and output, respectively. $\mathbf{x_t}$ is the input frame of the current time step.  $\mathbf{h_t}$ is the internal hidden state. $\mathbf{y_t}$ is the output, i.e., perception token. 

Benefit from the strong temporal modeling capability of  Selective SSM, our EPFE with only 56\,M parameters can learn long-term event-level spatiotemporal features with constant cost. As Appendix Fig.\,1 illustrates, perception tokens generated by EPFE can distinguish different events and noise frames, while maintaining a state of the event. Even after unrelated noise frames, the perception tokens are capable of refocusing on the event.

\subsection{Cognition Gate}
\label{sec:Cognition Gate}
\textbf{Motivation:} Cognition Gate needs to determine whether to invoke LLM in real-time, based on the current perception token and user query. For example, during the soccer match, some user query may only care LLM response for scoring moments, while others may need detailed tactical analysis.   

However, we find it highly challenging. For real-time consideration, we first employ a simple Transformer block with cross-attention to associate the user query with the visual encoder’s output for trigger decisions. However, it struggles to make accurate judgments at the right time step, as shown in Tab.\ref{tab:arch gate}. We identify two key reasons for this: 1) The Cognition Gate only has access to information up to the current step, lacking a global temporal perspective for decision-making. 2) User queries may not directly correspond to specific video frames in a retrieval-like manner (e.g., guide me to repair a bicycle or cooking).  

Therefore, the Cognition Gate requires the world knowledge of LLMs. However, due to the size of LLMs, even processing a single perception token as input incurs a high computational cost. Notably, the Gate only needs to generate a response/silence token for a single iteration.

% the deeper layers—primarily responsible for nuanced language generation—are unnecessary.

To achieve it, We propose the Shallow Layer Transfer method, which initializes the Gate using the early layers of the LLM and then fine-tunes it in a supervised autoregressive manner with the StreamVD dataset. As shown in Sec.~\ref{ablation:cognition gate}, we conducted extensive experiments to analyze its impact. The results indicate that the first few layers exhibit slight variations in behavior, allowing for scalability based on requirements. In this work, we use four layers.

\subsection{The Training Strategy of \sys}
\paragraph{Model Architecture}
%As illustrated in Figure \ref{fig:framework}, for each frame input to the model, we first utilize CLIP~\cite{radford2021learning} as a pre-trained, high-capacity feature extractor to extract general-purpose spatial features. These features are then processed by the EPFE to capture spatiotemporal dependencies and construct the perception token. Subsequently, the Cognition Gate evaluates whether to trigger the LLM to complete the cognition stage. To achieve this, it is necessary to train the EPFE, Cognition Gate, and LLM components, as shown in Figure \ref{fig:train frame}.

\begin{figure}[h] %H为当前位置，!htb为忽略美学标准，htbp为浮动图形
\centering %图片居中
\includegraphics[width=0.48\textwidth]{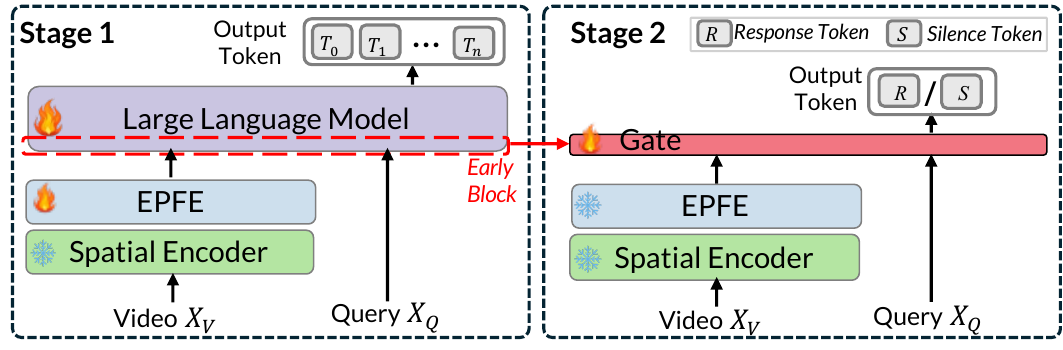} %插入图片，[]中设置图片大小，{}中是图片文件名
\caption{The two-stage training process of \sys. In the first stage, the EPFE spatiotemporal extractor and the LLM are jointly trained to align their representations. In the second stage, the Cognition module is trained to generate response/silence tokens. Both stages follow a supervised autoregressive training approach for language modeling. }%The training strategy of \sys. \sys adopts a two-stage training approach. First, it jointly trains the LLM and EPFE using existing video frames and captions, ensuring spatiotemporal feature alignment with the LLM's semantic understanding. After this pretraining phase, the trained LLM is used to initialize the Cognition Gate following the \textit{EarlyBlock} strategy. Finally, following the method described in Algorithm.\ref{alg:data organization}, data is prepared to train the Cognition Gate, enabling it to make informed decisions on whether to generate a response. } %最终文档中希望显示的图片标题
\label{fig:train frame} %用于文内引用的标签
\end{figure}

As illustrated in Figure \ref{fig:framework}, the \sys video LLM integrates CLIP~\cite{radford2021learning}, EPFE, Cognition Gate, and the LLM. CLIP is the pre-trained spatial feature extractor. EPFE then takes the spatial features as input as well as its internal state to extract spatiotemporal features and generate a perception token. Cognition Gate then evaluates the current perception token and user query to determine whether to invoke LLM. To achieve this, it is needed to train EPFE, Cognition Gate, and also align the representation space of EPFE and LLM, as shown in Fig.~\ref{fig:train frame}.    

% For each frame input to the model, we first utilize CLIP as a pre-trained, powerful feature extractor to extract general-purpose spatial features. These features are then processed by the EPFE to capture spatiotemporal dependencies and construct the perception token, as illustrated in Figure \ref{fig:framework}. Finally, the Cognition Gate evaluates whether to activate the LLM and update the Historical Synopsis based on the accumulated Immediate Memory. To achieve this, it is necessary to train the EPFE, Cognition Gate, and LLM components, as shown in figure \ref{fig:train frame}.

% Specifically, we employ CLIP [43] for initial spatial feature extraction. CLIP, as one of the most robust vision encoders, is widely adopted as the vision encoder in mainstream video foundation models [8, 38]. Subsequently, a linear layer combined with a state-assistance mechanism serves as the temporal-aware extractor to reduce redundancy in the general-purpose spatial features while maintaining temporal continuity in the video stream. 

\begin{algorithm}[h]
\caption{Streaming Video Dataset Construction}
\label{alg:data organization}
\begin{algorithmic}[1]
\REQUIRE Offline video dataset $C = \{c_i, t_i\}_{i=1}^n$ ($c_i$: caption at time $t_i$). 
\ENSURE Dataset $D$ with captions meeting StreamVD requirements and Cognition Gate labels.

\STATE \textbf{Step 1: Preprocessing}
\STATE $D \gets []$, $cap \gets c_1$, $time \gets t_s$
\FOR{$i = 2$ to $n$}
    \IF{$c_i \neq cap$}
        \STATE Append $(cap, time)$ to $D$
        \STATE $cap \gets c_i$, $time \gets t_i$
    \ENDIF
\ENDFOR
\STATE Append $(cap, time)$ to $D$

\STATE \textbf{Step 2: Silence-Response Labeling}
% \FOR{each $(d_i, d_{i+1}) \in D$}
%     \STATE Insert \texttt{</skip>} tokens between frames
 \FOR{each frame $f$ between $d_i$ and $d_{i+1}$}
    \STATE Label $f$ with \texttt{</response>} if captioned, else \texttt{</silence>}
    %\ENDFOR
\ENDFOR

\RETURN $D$
\end{algorithmic}
\end{algorithm}

\paragraph{Data Preparation}
\label{sec:data preparation}
Streaming datasets are constructed by processing existing offline datasets. StreamingVD tasks require the model to response as early as possible when relevant events occur while avoid redundant responses. 
%When performing StreamingVD, we aim for the AI assistant to respond as early as possible while avoiding redundant responses. 
To achieve this, we eliminate repetitive captions from the offline datasets and transform them into streaming datasets using the following methods, detailed in Algorithm \ref{alg:data organization}.
\begin{itemize}
    \item Preprocessing: Adjacent identical captions are merged, with the timestamp of the first occurrence being recorded as the annotation point for the caption.
    
    \item Silence-Response Labeling: To train the Cognition Gate, we insert $</silence>$ tokens into video frames between two adjacent captions. These tokens act as indicators to guide the Cognition Gate in generating the $</silence>$. The initial frame corresponding to captioned events is labeled with the $</response>$ token.
\end{itemize}

\paragraph{Training Strategy} We adopt a two-stage training strategy, as shown in Fig.~\ref{fig:train frame}. In contrast to previous typical two-stage training approaches \cite{li2023videochat,maaz2023video,zhang2023video}, 
our method uses a dedicated stage for Cognition Gate.  In the first stage, the LLM and EPFE are jointly trained using video frames and captions from the streaming dataset, ensuring spatiotemporal feature alignment with the LLM. The second stage trains only the Cognition Gate, which is initialized from the shallow layers of the LLM. This training also follows the LLM autoregressive manner to generate $</silence>$ or $</response>$ tokens, determining whether to invoke the LLM. Since the response label is much sparser than silence (e.g. 310:1), we introduce a balancing weight into the standard Cross Entropy loss during training to balance the labels. Refer to Sec.~\ref{sec:experiment_setting} and Appendix Sec.\,1 for more training details. 
%our method adopts a novel strategy by introducing a dedicated stage for information sufficiency assessment and decision-making. Specifically, the first stage, joint feature extraction, trains the EPFE and LLM collaboratively to process video frames, ensuring spatiotemporal feature alignment with the LLM's semantic understanding. The second stage, decision optimization, focuses on training the Cognition Gate to evaluate the sufficiency of perception token and make strategic decisions on whether to invoke the LLM. This additional stage enhances the model's ability to handle streaming video data with improved temporal reasoning and adaptive response capabilities.

\section{Experiment}

\begin{table*}[h] \fontsize{8.5}{12.5}\selectfont 
\caption{Online Experiments Results.}
\label{tab:online exper1}
\renewcommand\tabcolsep{5.0pt}
\centering
\begin{tabular}{c|c|cc|cccc|c}
\toprule[1.2pt]
\multirow{2}{*}{DataSet} & \multirow{2}{*}{Method} & \multicolumn{2}{c|}{Timing Alignment $\uparrow$} & \multicolumn{4}{c|}{Language Modeling Capability $\uparrow$} & \multirow{2}{*}{Training Cost} \\ \cline{3-8}
                         &        & TriggerAcc  &TimVal           & BLUE-1   & BLUE-4   & METEOR  & ROUGE-L    &                                \\ \hline
\multirow{3}{*}{Ego4d}   & VideoLLM-Online                                    & 32.34\%     &29.66\%    & 66.01     & 35.25        &31.12        &63.06                & 24h                                \\
                         & VideoLLM-MoD                          & 32.36\%    & 29.65\%      & 65.34     & 35.21        & 30.65        &63.02          &  14h                               \\
                         &\baseline Ours            &\baseline 43.34\% &\baseline39.73\%                &\baseline  67.12          &\baseline 39.26           &\baseline31.60           &\baseline65.71               &  (11+8)h                            \\ \hline
\multirow{3}{*}{Soccer}  &  VideoLLM-Online      & 31.25\%   & 28.34\%     & 75.36        &64.23        &50.92        &81.57                & 12h                                \\
                         &  VideoLLM-MoD        & 31.24\%          & 28.12\%           &74.96        &64.18        &  50.24     & 81.59             & 7h                                 \\
                         
                         &\baseline Ours       &\baseline 52.18\%  &\baseline 47.36\%         & \baseline 82.78      & \baseline66.70       & \baseline51.43       & \baseline82.04              &  (5+3)h                               \\
                         
                         \bottomrule[1.2pt]
\end{tabular}
\end{table*}

\begin{table}[h] \fontsize{7.5}{10}\selectfont 
\caption{Online Experiments Results. LION-FS was released on arXiv just two days before our submission, but its code was not released. Therefore, we could only obtain the Ego4D results based on its paper. }
\label{tab:online exper2}
\renewcommand\tabcolsep{2pt}
\centering
\begin{tabular}{c|c|cccc}
\toprule[1.2pt]
DataSet                 & Method & TimeDiff $\downarrow$ & Fluency $\uparrow$ & PPL $\downarrow$& Correctiness $\uparrow$ \\ \hline
\multirow{4}{*}{Ego4d}  &  VideoLLM-Online      &  2.04        & 45.0\%        &2.43         &48.1\%                 \\
                        & VideoLLM-MOD       & 2.04         &45.2\%         &2.41         &48.9\%                 \\
                        & LION-FS       & 2.15         &46.1\%         &2.09         &52.4\%                 \\
                        & \baseline Ours       &\baseline 1.89         & \baseline 60.2\%        &\baseline 2.02       & \baseline77.3\%                 \\ \hline
\multirow{3}{*}{Soccer} &VideoLLM-Online       &   15.62       & 46.35\%        &1.79         &53.5\%                 \\
                        &VideoLLM-MOD      &  15.68        & 45.34\%    & 1.80        & 53.3\%                 \\
                        & \baseline Ours       &  \baseline 14.02       & \baseline 70.35\%         & \baseline 1.59        & \baseline 89.2\%              \\ 
\bottomrule[1.2pt]
\end{tabular}
\end{table}

 \begin{table}[h] \fontsize{7}{8}\selectfont 
\caption{Results on COIN benchmarks(left to right): step recognition, task recognition,
 next forecasting, procedure forecasting, procedure forecasting with a goal. }
\label{tab:coin}
\renewcommand\tabcolsep{4pt}
\centering
\begin{tabular}{c|c|ccccc}
\toprule
\multirow{2}{*}{Method} & \multirow{2}{*}{\begin{tabular}[c]{@{}c@{}}Not use \\ HT 100M\end{tabular}} & \multicolumn{5}{c}{COIN Benchmark Top-1 Accuracy $\uparrow$} \\ 
                        &                                                                             & Step    & Task    & Next    & Proc    & Proc.+    \\ \hline
     ClipBERT    &  \checkmark   &     30.8     &  65.4  &-&-&-           \\ 
     TimeSformer    & $\times$     &46.5 &85.3 &34.0& 17.0 &40.1         \\ 
     Paprika    &    $\times$     & 51.0 &85.8 &43.2 &-&-         \\ 
     DistantSup    &  $\times$     & 54.1 &90.0 &39.4  &-&-       \\ 
     VideoTF    & $\times$        & 56.5 &91.0 &42.4& 40.2 &46.4         \\ 
     ProcedureVRL    &  $\times$       & 56.9 &90.8 &46.8 &-&-         \\ 
     VideoTaskGraph    &     $\times$    &  57.2& 90.5 &40.2   &-&-      \\ 
     VideoLLM-online-7B-v1        &  \checkmark  & 59.8 &92.1 &48.1& 47.9& 52.9         \\ 
     VideoLLM-online -8B-v1+        &  \checkmark  &  63.1 &92.7& 49.1 &49.8& 54.1         \\ 
     VideoLLM-MOD    &   \checkmark      & 63.4 &92.8 &49.7& 49.8 &53.3        \\ 
     \baseline Ours    & \baseline \checkmark   &  \baseline63.7  & \baseline93.2 &\baseline49.9 &\baseline49.8 &\baseline{54.2}      \\ 
\bottomrule
\end{tabular}
\end{table}

\begin{table}[h] \fontsize{7}{8}\selectfont 
\caption{Results on Ego4DLTA benchmark, evaluated on public server.
 ED@Z=20 denotes editing distance for future 20 actions. }
\label{tab:egolta}
\renewcommand\tabcolsep{3.5pt}
\centering
\begin{tabular}{c|c|c|ccc}
\toprule
\multirow{2}{*}{Method} & \multirow{2}{*}{\begin{tabular}[c]{@{}c@{}}Not use \\ Ego VLP\end{tabular}} 
 & \multirow{2}{*}{\begin{tabular}[c]{@{}c@{}}End-to \\ -End?\end{tabular}}& \multicolumn{3}{c}{Ego4D LTA ED@Z=20} \\ \cline{4-6} 
        &    & & Verb $\downarrow$   & Noun $\downarrow$   & Action $\downarrow$       \\ \hline
  CLIP      & \checkmark   & \checkmark &0.739& 0.769 &0.941                    \\ 
  EgoT2      & \checkmark   & \checkmark &0.722& 0.764 &0.935                  \\ 
  I-CVAE       & \checkmark   & \checkmark &0.753 &0.749 &0.931                   \\ 
  HierVL      & \checkmark   & \checkmark  &0.724 &0.735& 0.928            \\ 
  VideoLLM      & $\times$   &\checkmark &0.721& 0.725 &0.921                  \\ 
 VideoLLM-online-7B-v1       & \checkmark   & \checkmark & 0.697& 0.698& 0.897                   \\ 
 VideoLLM-online-8B-v1+       & \checkmark   & \checkmark &0.689 &0.671& 0.884                    \\ 
  VideoLLM-MOD       & \checkmark   & \checkmark &  0.689&0.676& 0.884                 \\ 
  \baseline Ours       &\baseline \checkmark   & \baseline \checkmark  &\baseline0.689 &\baseline0.655 &\baseline0.881                 \\ 
\bottomrule
\end{tabular}
\end{table}

\subsection{Experimental Settings}
\label{sec:experiment_setting}
\paragraph{Baseline and Datasets}
Since existing works focusing on streamingVD~\cite{chen2024videollm,wu2025videollm,huang2024online}, except for VideoLLM-Online~\cite{chen2024videollm}, do not provide open-source data or training methods, to demonstrate our outstanding performance, we re-implemented all previous works (VideoLLM-Online~\cite{chen2024videollm} and VideoLLM-MOD~\cite{wu2025videollm}) in the StreamingVD domain as baselines, using consistent settings across all experiments.  

For dataset selection, since the Ego4D dataset~\cite{grauman2022ego4d} is inherently collected in a streaming manner, containing various queries, real-time responses, and narrations, and has been used as an main evaluation dataset in works~\cite{wu2025videollm,chen2024videollm,li2025lion}, we selected it as one of our evaluation datasets.  Furthermore, to further demonstrate the effectiveness of our approach on long videos, we also conducted experiments on the SoccerNet-Caption dataset~\cite{giancola2018soccernet}. This dataset includes 471 complete soccer match videos with a total duration of 715 hours, where each video is approximately 45 minutes long and provides commentary at appropriate timestamps.

% The . We leverage this dataset to construct our streaming video dataset, specifically designed to train and evaluate our model components.

% For a fair comparison, we first evaluated our method on the Stream Ego4D , following the approach of VideoLLM-Online, which utilizes dense Ego4D timestamp-narrations to create a streaming dataset aimed at generating timely narrations akin to those produced by human annotators. To further demonstrate the effectiveness of our approach on long videos, we also conducted experiments on the SoccerNet-Caption dataset. This dataset includes 471 complete soccer match videos with a total duration of 715 hours, where each video is approximately 45 minutes long and provides commentary at appropriate timestamps.

\vspace{-3mm}
\paragraph{Implementation Details:}
Model training was conducted on videos sampled at 2\,fps, using 8 NVIDIA A100 GPUs. Each stage of the training process was conducted for one epoch. To optimize the learning rate, we applied the CosineAnnealingLR scheduler, with learning rates set to $2e-5$ and $2e-6$ respectively. Additionally, specific techniques were applied to the design of the Cognition Gate and methods to address the imbalance in the Silence-Response ratio, which are detailed in Section \ref{sec:ablation}.

\subsection{Evaluate Metrics}
To evaluate the effectiveness of \sys, we focus on assessing its \textbf{Timing Alignment} Capability and \textbf{Language Modeling} Capability. Existing evaluation metrics include LM-PPL (Language Modeling Perplexity) and LM-Correctness, which evaluate language modeling accuracy at specific timestamps. Additionally, the Time Difference (TimeDiff) metric is used to measure the temporal alignment of responses in an streaming video LLM, while the Fluency metric aims to provide a holistic evaluation of both linguistic quality and temporal consistency.

However, these metrics have several limitations:
\begin{itemize}
\item LM-PPL and LM-Correctness focus solely on the exact alignment between model outputs and reference narrations, ignoring aspects such as synonym substitutions, sentence fluency, and the ability to capture critical information.
\item TimeDiff, as a metric for timing alignment Capability, evaluates temporal alignment only within a single dialogue turn. It does not assess the Cognition Gate’s overall decision-making across the entire video stream.

% or its preferences for silence versus response generation.
\end{itemize}
To address these limitations, we introduce the following metrics: 
%\ting{can you add formulas to the metrics. it is very hard to understand the exact meaning without formulas.}
\begin{itemize}
    \item Trigger Accuracy (TriggerAcc): To evaluate whether the streaming video LLM responds at the correct time steps during entire streaming video.
    % Assesses the Cognition Gate's capability to identify moments where a response is necessary, emphasizing its effectiveness in capturing critical events during entire streaming video.
    \item Timing Validity (TimVal): To comprehensively evaluate whether the streaming video LLM consistently makes the correct decisions throughout the streaming video—speaking when necessary and remaining silent when appropriate.

    % Overall Perceptual Accuracy (TimVal): Measures the Cognition Gate's decision accuracy on a frame-by-frame basis across the entire video, providing a comprehensive evaluation of its judgment capability.
    \item BLUE~\cite{papineni2002bleu}, METEOR~\cite{banerjee2005meteor}, ROUGE-L~\cite{lin2004rouge}: Widely used evaluation metrics in the field of text generation~\cite{chang2024survey,liu2023g,chen2024self} and image narration~\cite{rao2024matchtime,pont2020connecting,changpinyo2021conceptual}, which fully consider factors such as synonym matching, stemming, word order, and key semantic expressions. These metrics enable a comprehensive assessment of dialogues in the StreamingVD process.
\end{itemize}

\subsection{Comprehensive Evaluation on Streaming and Offline Video}

\paragraph{Online Experiments:} We compared our model with all existing streaming dialogue models~\cite{wu2025videollm,chen2024videollm,li2025lion} in Tab \ref{tab:online exper1} and Tab \ref{tab:online exper2}. Under comparable training costs, experiments on real-time Ego4D and SoccerNet demonstrate superior performance across all metrics. Additionally, \sys can process video streams at a 10× higher FPS. This improvement is attributed to the event-level modeling capability of EPFE and the strong timing alignment capability of the cognition gate in the perception stage, which together provide the LLM in the cognition phase with a concise yet comprehensive representation.
\vspace{-3mm}
\paragraph{Offline Experiments:} 
We also leveraged the COIN~\cite{tang2019coin}  and Ego4D LTA~\cite{huang2023palm} benchmark to demonstrate \sys effectiveness in traditional offline video scenarios, covering six common benchmarks: step recognition, step forecasting, task summarization, procedure forecasting, and procedure forecasting with a goal. As shown in Tab \ref{tab:coin} and Tab \ref{tab:egolta}, our approach achieves state-of-the-art performance across all tasks.

% We also demonstrate our proposed \sys can perform well on traditional offline video scenarios, including recognition, summarization, and forecasting tasks. As shown in Tab \ref{tab:coin} and Tab \ref{tab:egolta}, Ours achieves state-of-the-art performances in COIN and Ego4D LTA benchmarks.

%  \subsection{Understanding Learned Knowledge in Perception Token}
% In this section, we conducted a visual examination of the effectiveness of perception tokens as representations of the current temporal context. As streaming video inputs continue, we preserved all perception tokens perceived by the cognition gate between two consecutive cognition phases. We then calculated the cosine similarity between these tokens. The results indicate that perception tokens can effectively differentiate between relevant and irrelevant events occurring within the current situation while maintaining a strong memory of the main event. Even after encountering unrelated events, perception tokens are capable of refocusing on the primary event. Moreover, throughout the entirety of an event, perception tokens exhibit robust memory retention, maintaining a high similarity with earlier phases of the event even in its later stages.
% \begin{figure}[h] %H为当前位置，!htb为忽略美学标准，htbp为浮动图形
% \centering %图片居中
% \includegraphics[width=0.4\textwidth]{figure/visualization_two_event.pdf} %插入图片，[]中设置图片大小，{}中是图片文件名
% \caption{A cosine similarity heatmap of frames across two consecutive events.} %最终文档中希望显示的图片标题
% \label{fig:com time} %用于文内引用的标签
% \end{figure}

\subsection{Real-time Inference Efficiency}
Standard frame rates for films are typically 24 FPS, which is considered the minimum requirement for smooth motion perception. In contrast, games often operate at much higher frame rates, ranging from 30 FPS to over 100 FPS, to ensure a immersive experience. To validate the real-time performance of \sys, we conducted inference efficiency evaluations on NVIDIA A100 and H100 GPUs.

\begin{figure}[h] %H为当前位置，!htb为忽略美学标准，htbp为浮动图形
\centering %图片居中
\includegraphics[width=0.4\textwidth]{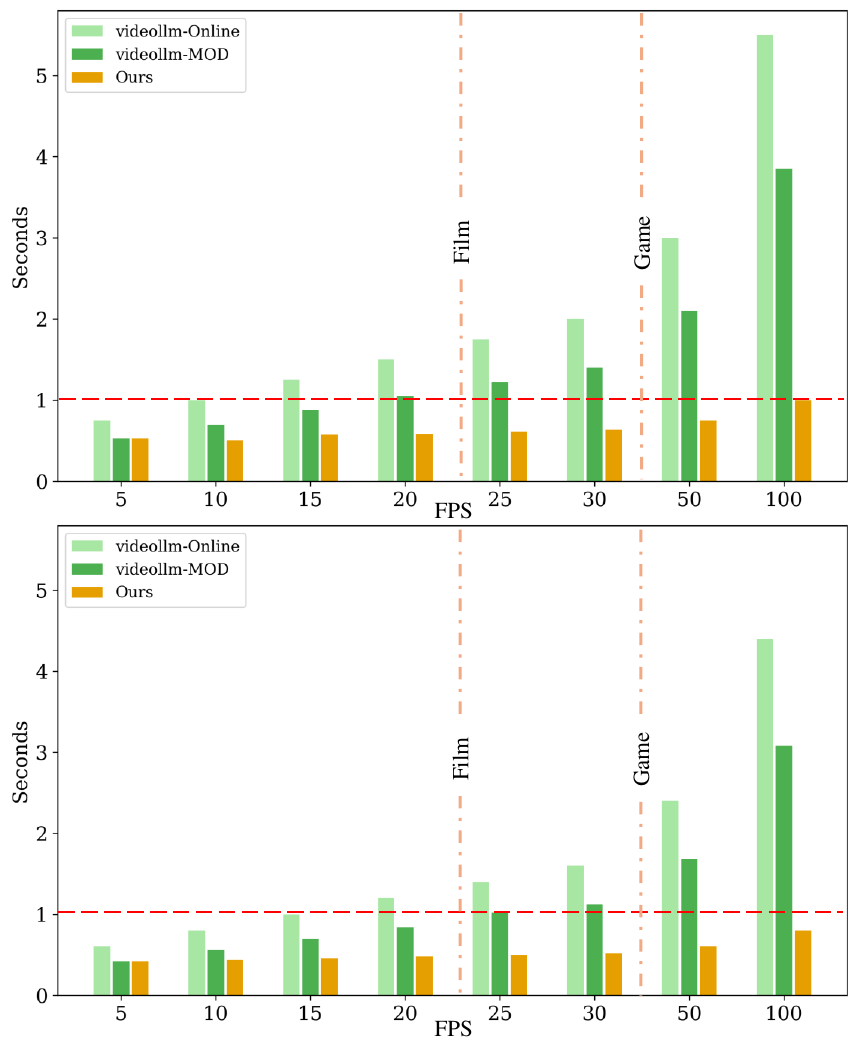} %插入图片，[]中设置图片大小，{}中是图片文件名
\caption{Comparison of running time for processing 1 second of streaming video frames using different methods on A100 (Top) and H100 (Bottom).} %最终文档中希望显示的图片标题
\vspace{-1mm}
\label{fig:com time} %用于文内引用的标签
\end{figure}

Specifically, we compared VideoLLM-Online, VideoLLM-MOD, and \sys by measuring their actual processing time for 1-second video across input frame rates from 5 to 100 FPS. As shown in Figure \ref{fig:com time}, a processing time under 1 second indicates real-time performance without disrupting playback.

The results reveal that VideoLLM-Online and VideoLLM-MOD struggle to process video streams beyond 10 FPS, leading to potential latency issues in real-world applications. In contrast, our method achieves true real-time performance, efficiently handling both standard film/TV frame rates and high-refresh gaming scenarios up to 100 FPS, ensuring real-time responses.

\section{Perception Phase Visualization Experiment}

\begin{figure*}[t] %H为当前位置，!htb为忽略美学标准，htbp为浮动图形
\centering %图片居中
\includegraphics[width=1\textwidth]{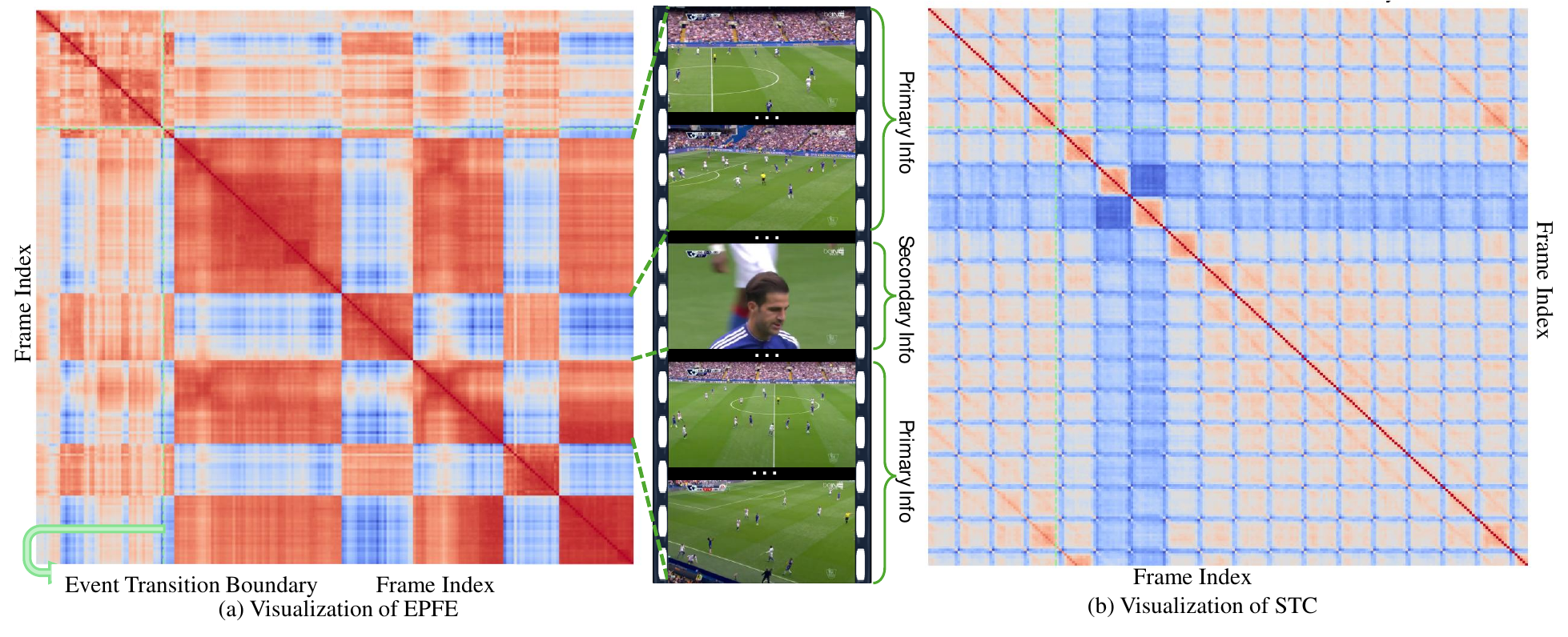} %插入图片，[]中设置图片大小，{}中是图片文件名
\caption{A cosine similarity heatmap of frames across two consecutive events.} %最终文档中希望显示的图片标题
\label{fig:com visual} %用于文内引用的标签
\end{figure*}

To visualize the perception phase, we conducted the following experiment: As streaming video inputs continued, we preserved all perception tokens generated by the Cognition Gate between two consecutive cognition phases. We then computed the cosine similarity between these tokens to analyze their consistency over time.

The results indicate that perception tokens generated by EPFE can effectively distinguish between relevant and irrelevant events while maintaining a strong memory of the primary event. Even after encountering unrelated events, the perception tokens are capable of refocusing on the main event. Furthermore, throughout the entire event, perception tokens exhibit robust memory retention, maintaining a high similarity with earlier stages of the event even in later phases, as shown in figure.\ref{fig:com visual}.(a).

We repeated this experiment on STC, as shown in figure \ref{fig:com visual}.(b), but its perception tokens failed to maintain long-term feature similarity, capturing only local spatiotemporal features. This further highlights the superiority of EPFE in preserving event-level semantics over extended video sequences.

\section{Ablation}
\label{sec:ablation}
\subsection{Impact of Silence-Response Sample Imbalance}
% In this section, we explored the imbalance in the Silence-Response ratio within the dataset and derived an empirical balancing formula through experiments. For detailed experimental results, please refer to the supplementary material.

As described in the data preparation process, during the Silence-Response Labeling stage, the dataset inherently exhibits a severe imbalance in the Silence-Response ratio. For example, in a 1-minute video at 30 FPS, if only 5 frames require a response, the silence-response ratio becomes 360:1. Such an extreme imbalance can significantly impact the training process. 

To mitigate this issue, we introduce a balancing weight $W_s$ into the standard Cross Entropy (CE) loss during training, where $W_s$ and $1 - W_s$ represent the weights for silence and response, respectively. We conducted extensive ablation studies on Silence-Response sample balancing based on the weight $ W_s $. The results indicate that, compared to the standard cross-entropy loss, $ W_s $ significantly improves the final performance. However, an unfortunate observation is that the optimal $ W_s $ varies considerably, even differing by orders of magnitude. To address this, we analyzed the ratio $ P $ of $ </silence> $ to $ </response> $ tokens in the Ego4D and SoccerNet datasets, which are $310:1$ and $71:1$, respectively. Based on these observations, we derive an empirical formula for the optimal balancing weight $W_s^{\text{opt}}$:  
\begin{equation}
W_s^{\text{opt}} \approx  10 P
\end{equation}

\begin{table}[h] \fontsize{8}{9}\selectfont 
\caption{Perceptive Judgement Capability}
\renewcommand\tabcolsep{4pt}
\centering
\begin{tabular}{c|c|ccc}
\toprule[1.2pt]
Datasets & Method     & TimeDiff $\downarrow$ & TriggerAcc $\uparrow$ & TimVal $\uparrow$\\ \hline
\multirow{6}{*}{Ego4D} &Standard CE & 2.05 & 31.37\% &30.91\%     \\
&0.10 & 1.92  &41.34\%  & 37.60\%   \\
&0.12 & 1.88  &42.16\%  & 38.63\%   \\
&\baseline0.15 & \baseline1.89 &\baseline 43.34\% &\baseline 39.73\%     \\
&0.17 & 1.95   & 40.23\%   &39.44\%  \\
&0.20 &  1.97    & 35.35\%   &34.60\% \\ 
\hline
\multirow{6}{*}{Soccer}
& Standard CE &15.89  &31.34\% &29.61\%  \\
& 0.01 &14.33& 50.35\% &41.32\%\\
& 0.02   &14.35  &51.62\% &44.8\%\\
&\baseline 0.03 &\baseline14.02  &\baseline52.18\%  &\baseline47.36\%   \\
& 0.04   &14.03 & 44.69\% & 42.37\%  \\
&0.05     &14.12 & 41.22\% & 39.82\%   \\
                         \bottomrule[1.2pt]
\end{tabular}
\end{table}

\subsection{The Ablation of Cognition Gate}
\label{ablation:cognition gate}
The Cognition Gate plays a crucial role throughout the perception-cognition interleaving process. We conducted detailed ablation experiments to analyze its impact.

\paragraph{The Architecture of Cognition Gate}
The structure of the gate directly determines its capabilities. To this end, we constructed multiple classification networks using MLP projectors, Cross-Attention, and Transformers.~\cite{vaswani2017attention} Additionally, we built a set of autoregressive networks using an LLM block. The results are shown in Tab.\ref{tab:arch gate}. It is evident that the LLM-like structure achieved better performance. For a detailed analysis, please refer to Section \ref{sec:Cognition Gate}.
\vspace{-0.5mm}
\begin{table}[h] \fontsize{8.5}{9}\selectfont 
\caption{Ablation Study on the architecture of cognition gate }
\vspace{-0.5mm}
\label{tab:arch gate}
\renewcommand\tabcolsep{4pt}
\centering
\begin{tabular}{c|cccc}
\toprule[1.2pt]
Architecture     & TimeDiff $\downarrow$  & TriggerAcc $\uparrow$ &TimVal $\uparrow$\\ \hline
  Linear Layer &  5.06    & 20.13\%   &17.65\%    \\
  MLP projector + Linear&  4.34    &21.75\% &18.33\%    \\
  Transformers + Linear&3.78      &21.58\%   &18.97\%    \\
  Cross-Attention + Linear&3.64      &24.34\%   &20.36\%    \\
  \baseline Single LLM block &\baseline  2.22 & \baseline35.35\% &\baseline 32.11\%     \\
                         \bottomrule[1.2pt]
\end{tabular}
\end{table}
\vspace{-3mm}

\paragraph{Initialization Strategies}
We investigate three initialization strategies for the Cognition Gate to enhance its effectiveness: (i) Random Initialization: All layers are initialized from a standard random distribution without utilizing prior knowledge from the LLM.
(ii) SkipBlock Initialization: The Cognition Gate is initialized by uniformly selecting layers from the LLM, with the number of skipped layers determined by the ratio of the Cognition Gate's depth to the total number of LLM layers. (iii) EarlyBlock Initialization: The first few layers of the LLM are directly used to initialize the Cognition Gate, leveraging the LLM's early-stage processing capabilities. 

% Experimental results indicate that EarlyBlock Initialization effectively balances adaptation and computational efficiency, making it a preferable choice. 

\begin{table}[h] \fontsize{9}{9}\selectfont 
\caption{Ablation Study on the Strategy of Initialization }
\label{tab:init}
\renewcommand\tabcolsep{3.5pt}
\centering
\begin{tabular}{c|cccc}
\toprule[1.2pt]
Strategy of Initialization     & TimeDiff $\downarrow$  & TriggerAcc $\uparrow$ &TimVal $\uparrow$\\ \hline
Random  &   2.10   & 39.65\%  &37.35\%    \\
SkipBlock  & 1.89  & 40.23\%  &37.67\%    \\
\baseline EarlyBlock & \baseline1.89 &\baseline 43.34\% &\baseline 39.73\%     \\
                         \bottomrule[1.2pt]
\end{tabular}
\end{table}

\paragraph{Number of Blocks}
The Cognition Gate consists of multiple layers, and its depth plays a crucial role in balancing computational efficiency and perceptual accuracy. 
\begin{table}[h] \fontsize{9}{9}\selectfont  
\caption{Ablation Study on the Strategy of Initialization }
\label{tab:num block}
\renewcommand\tabcolsep{7pt}
\centering
\begin{tabular}{c|ccc}
\toprule[1.2pt]
\#Num of Block     & TimeDiff $\downarrow$ & TriggerAcc $\uparrow$ &TimVal $\uparrow$   \\ \hline
2 & 1.92  & 41.53\% & 37.65\%    \\
3 & 1.90  & 42.35\% & 38.34\%    \\
\baseline 4 & \baseline1.89 &\baseline 43.34\% &\baseline 39.73\%    \\
5 & 1.88  & 42.67\%  & 38.56\%   \\
6 & 1.89  & 42.56\% & 38.34\%   \\
                         \bottomrule[1.2pt]
\end{tabular}
\end{table}

The results in Tab.\ref{tab:init} and Tab.\ref{tab:num block} show that fine-tuning the shallow layers of the LLM yields the best performance. We attribute this to the fact that gate operates within a limited decision space (binary: Yes/No), making deeper layers—primarily responsible for nuanced language generation—unnecessary. Thus, we propose the Shallow Layer Transfer method to construct the Cognition Gate, as detailed in Section~\ref{sec:Cognition Gate}.

\subsection{The performance of Event-preserving Feature extractor}
To evaluate the role of EPFE in the perception phase, we compare it with commonly used extractors in VideoLLMs, including: 1) The Q-Former~\cite{li2023blip}. 2) The Spatial-Temporal Convolution Connector (STC Connector)~\cite{cheng2024videollama}, which has demonstrated outstanding performance in videollm and has been widely adopted~\cite{zhang2023video,li2023videochat,wang2024lstp}. Notably, STC employs 3D convolution for spatial-temporal aggregation. To preserve local visual patterns during spatial compression, it follows work~\cite{cha2024honeybee} by incorporating a RegStage block~\cite{radosavovic2020designing} before and after the 3D convolution, which has been shown to enhance spatial understanding. Experimental results indicate that EPFE achieves superior performance.

\begin{table}[h] \fontsize{9}{9}\selectfont 
\caption{Perceptive Judgement Capability}
\renewcommand\tabcolsep{7pt}
\centering
\begin{tabular}{c|ccc}
\toprule[1.2pt]
 Method     & TimeDiff $\downarrow$ & TriggerAcc $\uparrow$ & TimVal $\uparrow$\\ \hline
Q-former &3.78  &26.65 \% &25.31\%     \\
STC   &3.56  &27,54\% &26.87\%   \\
EPFE  & \baseline1.89 &\baseline 43.34\% &\baseline 39.73\%     \\
                         \bottomrule[1.2pt]
\end{tabular}
\end{table}

% In the second stage of training, we explored the imbalance in the Silence-Response ratio within the dataset and derived an empirical balancing formula through experiments. For detailed experimental results, please refer to the supplementary material.

\section{Related Work}
\subsection{Offline VideoLLMs.}
% Recent advancements in large language models (LLMs) have driven the development of multimodal large language models (MLLMs), which integrate textual data with visual and other modalities. With the progress of the BLIP and LLAVA series in image-text models, researchers have begun expanding the scope of image data to include videos. 
In recent years, there has been a surge of interest in leveraging
Large Language Models (LLMs) for video understanding, leading to the development of offline videoLLMs. The biggest challenge for it is managing memory requirements~\cite{squire2015memory,cai2022memot,cheng2022xmem,song2024moviechat,he2024ma,fan2024videoagent} and compressing redundant frame features~\cite{zhao2023learning,jin2024video,sun2023fine,cheng2024videollama,huang2024lita}. LaViLa~\cite{liu2019learning} addresses this challenge by transforming LLMs into narrators that generate detailed descriptions of long videos from visual inputs. Other models, such as ChatVideo~\cite{wang2023chatvideo} and MM-VID~\cite{lin2023mm}, convert videos into text to improve comprehension. MovieChat~\cite{song2024moviechat}, on the other hand, combines all frame features using a simple averaging strategy. However, in the context of online streaming, there has been limited exploration into how VideoLLMs can meet the demands of temporal alignment and real-time processing for streaming video inputs.

\subsection{Online VideoLLMs.}
Recent studies have introduced VideoLLMs specifically designed for online stream understanding, while most existing works~\cite{distreaming,zhang2024flash,qian2025streaming,xiong2025streaming,zhou2024streaming} focus on improving offline video efficiency to meet real-time requirements. However, they fail to proactively decide when to respond based on user requirements. For example, ReKV~\cite{distreaming} implemented a sliding-window attention mechanism and a KV-cache system to reduce computational overhead. To address this, VideoLLM-Online~\cite{chen2024videollm} introduces per-step LLM invocation for online video dialogue, as shown in Fig.\ref{fig:paradigm}. However, its performance is constrained by limited per-frame processing capability due to the absence of effective frame compression strategies. VideoLLM-MOD~\cite{wu2025videollm} mitigates this issue by incorporating a mixture-of-depth approach for efficient visual token computation, enabling higher visual input resolutions. Building on MOD, LION-FS~\cite{li2025lion} proposes Token Aggregation Router and Token Dropping Router as the Fast Path, adaptively aggregating distinct features while discarding redundant ones. However, they remain limited to processing videos at only 10 FPS.

Overall, prior work remains restricted by the contradiction between sequential full-frame processing and the extremely high efficiency requirements of streaming video, lacking breakthrough paradigm designs explicitly suited for streaming frame inputs. To address these challenges, our approach introduces a novel perception-cognition interleaving paradigm and proposes the \sys framework, enabling VideoLLM to achieve proactive responses and real-time streaming video understanding, with a maximum throughput of up to 100 FPS.

\section{Visualization of demo}
\begin{figure*}[!htbp] %H为当前位置，!htb为忽略美学标准，htbp为浮动图形
\centering %图片居中
\includegraphics[width=0.9\textwidth]{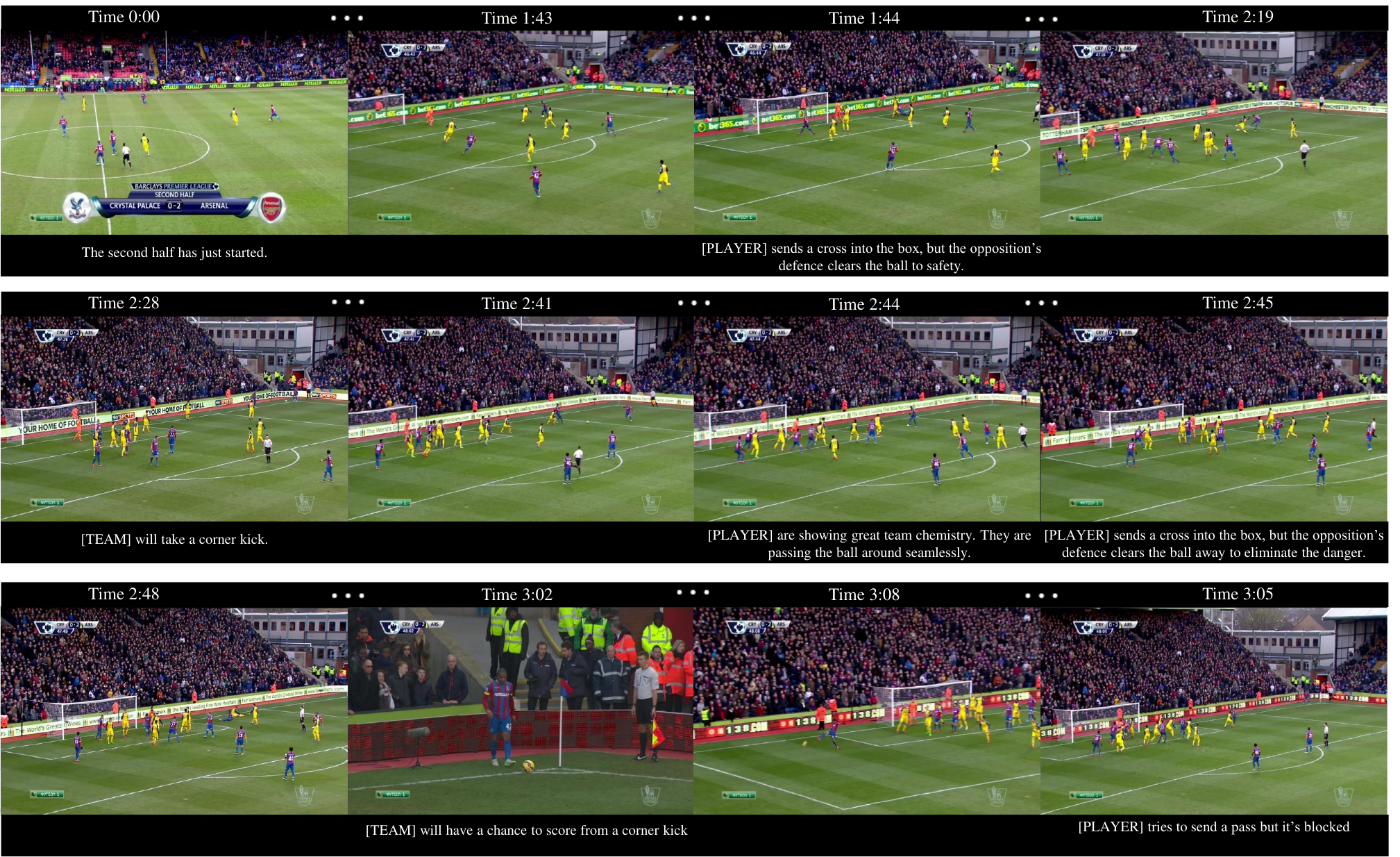} %插入图片，[]中设置图片大小，{}中是图片文件名
\caption{An illustrative snippet of StreamingVD on a live football match. The query at the beginning is: "Hey, Robot, can you watch the football game with me and provide commentary?"} %最终文档中希望显示的图片标题
\label{fig:demo} %用于文内引用的标签
\end{figure*}

we show an illustrative snippet of StreamingVD on a live football match. The query at the beginning is: "Hey, Robot, can you watch the football game with me and provide commentary?",as show in figure \ref{fig:demo}.

\section{Conclusion}
In this paper, we introduce a novel perception-cognition interleaving paradigm called \textbf{``event-gated LLM invocation''} and propose the \sys framework, addressing the contradiction between sequential full-frame processing and the extremely high efficiency requirements of streaming video. As a result, \sys can process video at up to 100 FPS, achieving the SOTA performance across all existing metrics and benchmarks.

{
    \small
    \bibliographystyle{ieeenat_fullname}
    \bibliography{main}
}

% \newpage
% \newpage
% \newpage
% \newpage
\appendix

\end{document}